\def\year{2021}\relax
\documentclass[letterpaper]{article} 

\usepackage{aaai21}  

\usepackage{times}  
\usepackage{helvet} 
\usepackage{courier}  
\usepackage[hyphens]{url}  
\usepackage{graphicx} 
\usepackage{natbib}  
\usepackage{caption} 
\usepackage{color}
\usepackage{wrapfig}

\usepackage{appendix}
\usepackage{subfig}
\usepackage{chngpage}
\usepackage{booktabs}
\usepackage[ruled,vlined]{algorithm2e}
\usepackage{tabularx}

\graphicspath{ {./images/} }
\usepackage{amssymb,amsmath, amsthm}
\usepackage[switch]{lineno}
\begin{document}

\title{Learning Neural Networks on SVD Boosted Latent Spaces for Semantic Classification}
\author{ Sahil Sidheekh \\
}

\affiliations{
    Indian Institute of Technology, Ropar \\
    2017csb1104@iitrpr.ac.in
}

\maketitle

\begin{abstract}
 The availability of large amounts of data and compelling computation power have made deep learning models much popular for text classification and sentiment analysis. Deep neural networks have achieved competitive performance on the above tasks when trained on naive text representations such as word count, term frequency, and binary matrix embeddings. However, many of the above representations result in the input space having a dimension of the order of the vocabulary size, which is enormous. This leads to a blow-up in the number of parameters to be learned, and the computational cost becomes infeasible when scaling to domains that require retaining a colossal vocabulary. This work proposes using singular value decomposition to transform the high dimensional input space to a lower-dimensional latent space. We show that neural networks trained on this lower-dimensional space are not only able to retain performance while savoring significant reduction in the computational complexity but, in many situations,  also outperforms the classical neural networks trained on the native input space. 
\end{abstract}


\section{Introduction}

With the emergence of deep learning \cite{goodfellow2016deep}, artificial intelligence has taken off in all possible fields. Machines capable of analyzing and classifying natural language data \cite{deng2018deep,conneau2016very} has gained popularity in recent years. While efficient neural network models \cite{vaswani2017attention,devlin2019bert} have emerged to process natural language data, learning itself can be slow and computationally expensive, given the inherent structure of text data representations. Classical approaches for representing textual data consists of document-term matrices \cite{LACHENBRUCH2012xv} that indicate the presence of words from a known vocabulary in a given document. This is also known as a  bag of words (BoW) representation. More specifically, the $i^{th}$ row $j^{th}$ column of the document term matrix captures the presence of the $j^{th}$ word from a known vocabulary in the $i^{th}$ document. Given such a data representation one may then train machine learning models to make inference from the textual data.

Over the years, several different forms have been devised for the document-term matrix to serve as a better manipulable representation for text data. However all have the inherent structure of having dimensions of the order of the vocabulary size, which may be quite huge in many domains. This results in huge and often sparse representations for the text data. Training models, especially neural networks on such huge and redundant input spaces is inefficient as it causes an exponential blow up in the number of model parameters to be learned.
Perhaps a better way to learn would be to transform the text data to a lower dimensional latent space that can not only de-noise the data, but also retain necessary meaningful information to achieve considerable performance on the down stream tasks. In this work we propose to use singular value decomposition to project the data into the space defined by the top-$K$ principal components of the training data matrix. This latent space would then serve as the input space for a neural network to make inference in terms of natural language data classification. We show that neural networks trained on such latent spaces are able to achieve considerable performance while significantly reducing the computational and storage overhead.

\section{Literature Review}

\begin{figure*}
    \centering
    \includegraphics[width=0.6\linewidth]{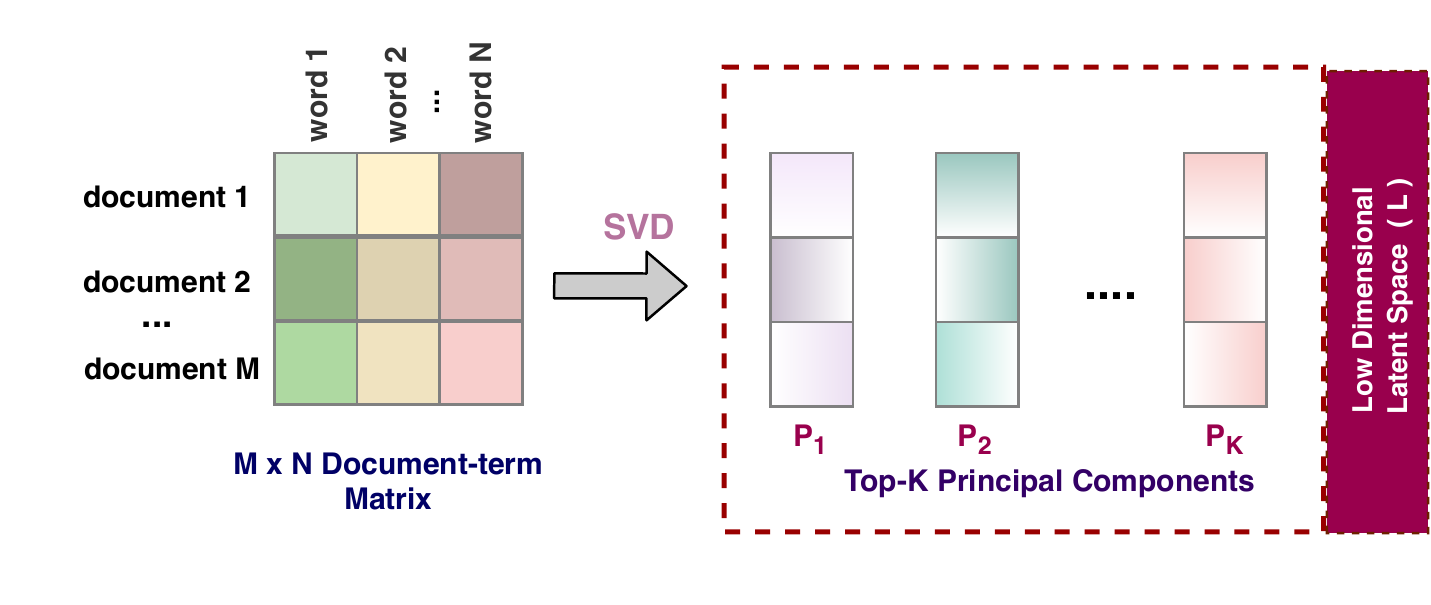}
    \\
    \caption{Defining a low dimensional latent space through the Top-K principal components of the document term matrix}
    \label{fig:latent_space}
\end{figure*}

Efficiently representing textual data is the fundamental task that needs to be addressed prior to devising sophisticated machine learning models that can make natural language inference. The N-gram representation as proposed by \cite{rajaraman2011mining} is one of the most popular methods for representing the text data, where the documents are represented as bag-of-words. Further studies have suggested various forms for the N-gram representation that considers the frequency of the words in the documents. The most succesfull in this direction has been the use of term-frequency weighed by the inverse document frequency, popularly known as TF-IDF \cite{ramosusing}. However, the N-gram representations have been criticized by many for its inefficacy over training machine learning models for text categorization \cite{kim2005dimension,silva2010distributed} owing to its inherent sparse and redundant structure. Thus, transforming the input space to lower dimensional latent space has gained attention in recent years.

Neural networks have been shown effective in automatically retrieving important features from input data that can outperform most of existing handcrafted feature representations in audio and visual domains. On similar lines, \cite{kim2014convolutional,sharif2014cnn} propose using a learnable embedding layer that will project the input text data representation to a learned latent space optimal for the down stream task. Though sophisticated, this approach requires learning of a large amount of additional parameters and requires the availability of huge amounts of training data, albeit which the model can easily overfit the training data distribution. 

The work by \cite{altszyler2016comparative} presents the efficiency of latent semantic analysis that utilizes SVD to obtain important word vectors in tasks pertaining semantic similarity. Developing over this, on similar lines with this work, \cite{8697314} studies the effect of SVD on training convolutional neural networks on TF-IDF vectors in terms of performance gains. Our work draws motivation from the above mentioned works and aim to study the ability of SVD to project the input data into an efficient latent space that significantly reduces the computational complexity and retains the performance of a feed-forward neural network for text classification. Unlike the more sophisticated word-embedding approaches that require training additional auxilary models, we deterministically project the data into the top $K$ principal components of the training data distribution and thus has no additional overhead of defining auxilary models or tuning additional hyperparameters.

\section{Methodology}

\subsection{Latent Space Transformation}

Let us denote by $C$ the document-term matrix of size $M \times N$. We define a latent space of dimension $K$ as the subspace spanned by the singular vectors of $C$. The existence of the above mentioned space is proven by the following well established theorem :

\newtheorem{thm}{Theorem} 
\begin{thm} \cite{wall2003singular}

Let $r$ be the rank of the document term matrix $C \in M_{m \times n}(R)$. Then, $\exists$ a decomposition $ C = UDV^{T}$, where 
    \begin{itemize}
        \item $U \in M_{m \times r}(R)$,  $V \in M_{n \times r}(R)$
        \item $D \in M_{r}(R)$ is a diagonal matrix such that the diagonal elements ($\sigma_i$) are the square root of the eigen values ($\lambda_i$) of $C^TC$.
    \end{itemize}
The diagonal elements ($\sigma_i$) are called the singular values of $C$ and the columns $\mathbf{u_i}$ of $U$ and $\mathbf{v_i}$ of $V$ are called the left and right singular vectors of $C$ respectively.
\end{thm}

The above decomposition can also be equivalently written as : $C = \sum_{i=1}^{r} \sigma_i   \mathbf{u_i}.\mathbf{v_i}^T$. Thus if we consider only the first $K$ terms in the above summation, we can get an equivalent $K$ dimensional subspace defined by the corresponding singular vectors. The reason as to why the above space should be the best $K$ dimensional subspace that is best representative of the original space is given by the Eckart-Young theorem. The following theorem formally summarizes this to fit our context : 

\begin{thm} 
Consider the document-term matrix $C \in M_{m \times n}(R)$ of rank $r$. Let $\mathbf{v_1}$,  $\mathbf{v_2}$, $\mathbf{v_3}$, ... $\mathbf{v_r}$, be the singular vectors of $C$ as discussed above. Then $ \forall k $ such that $1\leq k \leq r$, the subspace $V_k$ spanned by the first $K$ singular vectors  $\mathbf{v_1}$,  $\mathbf{v_2}$, $\mathbf{v_3}$, ... $\mathbf{v_k}$ of $C$ forms the best-fitting $K$ dimensional subspace for $C$.
\end{thm}

Thus projecting the document vectors onto to the subspace spanned by the top-$K$ singular vectors of $C$ gives an efficient $K$ dimensional representation that captures the maximum information pertaining the context and and topic embedded in the training data distribution. Such a projection, that retains information while reducing dimensionality implicitly helps reduce the noise and redundancy inherent in the input data. This latent space, thus is a proficient input space for training deep neural networks for downstream tasks such as text classification.

\subsection{Text Classification}

\begin{figure*}
\centering
    \includegraphics[width=0.8\linewidth]{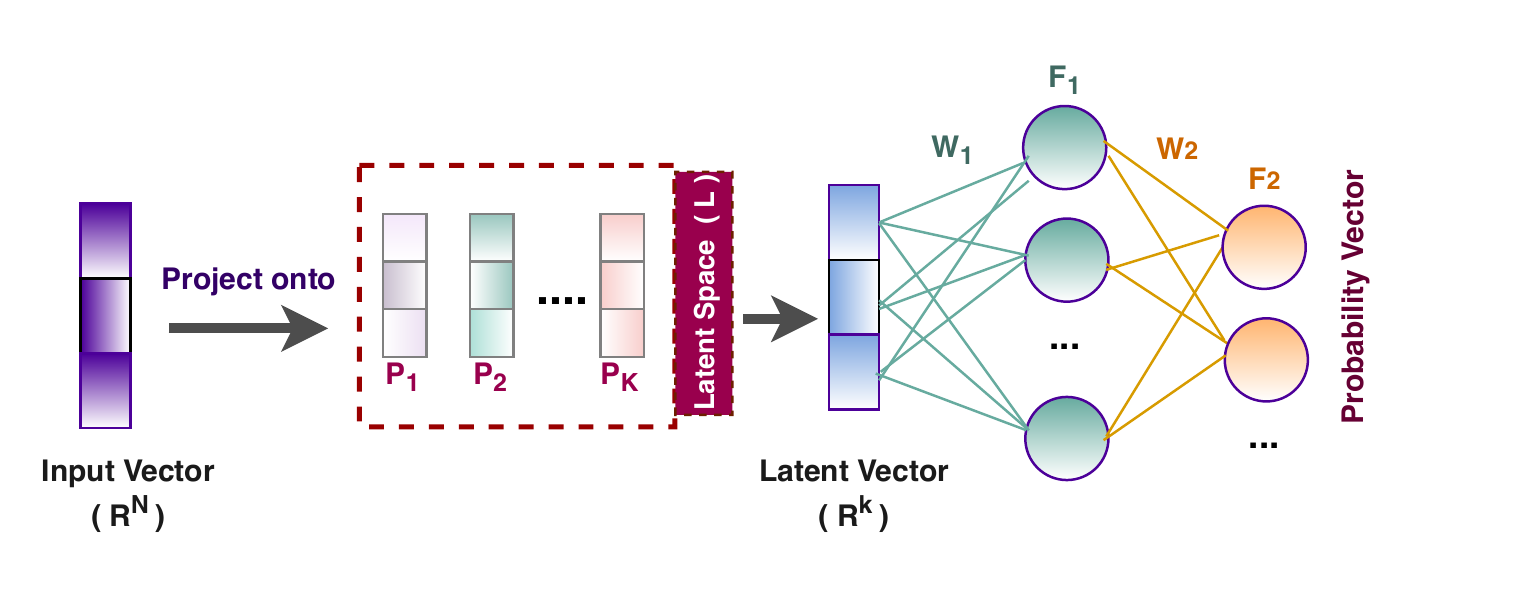}
    \centering
    \caption{Training a feed-forward neural network on the the SVD boosted latent space}
    \label{fig:model}
\end{figure*}

The ability to comprehend the context and topic from a given document is an essential trait of human intelligence. We have the ability to group the occurences of words in documents to categorize them under different domains. The latent space transformation defined earlier project data into the directions of maximum variance which can be visualized as the context and topics existent in the training data distribution. For a text classification task, if the above transformation results in a mutually non-overlapping distribution of the target classes, a mere clustering of the transformed document vectors will suffice to make inference on test data. However, as we will see in the experiments, the latent space transformation often results in mutually overlapping target class distributions, making inference a hard task. There exists some highly non-linear and complex mapping from the latent space of the document vectors to its target class.

\begin{thm} \cite{HORNIK1991251}
  A Feed-forward neural network of sufficient capacity is a universal function approximator.
\end{thm}

We can thus parameterically model the above mapping using a feed-forward neural network.

\subsubsection{Neural Network Formulation}
Let $G(\mathbf{x})$ represent the neural network where $\mathbf{x} \in R^k$ is the latent representation of the input document vector. We define a single hidden layer neural network with 128 nodes. The output of the neural network is a 2-dimensional probability vector, as we consider as binary classification task. Then we can formally define the neural network as : 
\begin{center}
$G(\mathbf{x}) = f_{2}(W_2^T(f_1(W_1^T(\mathbf{x})+ b_1)+b_2)$
\end{center}

where, $W_1 \in R^{ k \times 128}$ represents the weights connecting the input to the hidden layer, $W_2 \in R^{ 128 \times 2}$ represents the weights connecting the hidden layer to the output layer, $b_1 \in R^{128}$ and $b_2 \in R^2$ are the corresponding bias vectors and $f_1$ and $f_2$ are non linear activation functions applied element-wise (for each $x_i$ in $\mathbf{x}$) and defined by, 
\begin{itemize}
    \item $f_1(x_i) = max ( 0 , x_i ) $, is the rectified linear unit (ReLU) activation layer
    \item $f_2(x_i) = \underset{j \neq i}{\sum} e^{x_i}/e^{x_j} $, is the softmax activation layer 
\end{itemize}

We can easily observe that the dimension of $W_1$ increases with the dimension of the input vector $\mathbf{x}$ multiplied by the number of nodes in the first layer. For a vocabulary of size N and \#hidden nodes 100, the parameters to be learned in the first layer is 100N. This leads to an exponential blow up in the computational complexity for large vocabulary and motivates the need for an efficient low dimensional latent space to train the neural network. 
\subsubsection{Neural Network Training}

Given the neural network $G$ parameterized by the weights $W_1$ and $W_2$, we can learn the mapping from the latent space to the target class by minimizing the divergence between the target output and the predicted output given by the neural network. Thus, given the latent document vectors $\mathbf{x}$ and their corresponding target class $\mathbf{y}$ and the divergence $DIV(G(\mathbf{x}),\mathbf{y})$, training the neural network boils down to the below function optimization:
\begin{center}
    $ \underset{W_1,W_2}{min} DIV(G_{W_1,W_2}(\mathbf{x}),y)$
\end{center}
This minimization problem can be solved using gradient descent as the divergence is differentiable with respect to the parameters of the neural network. On convergence to the minima ($W_1^*,W_2^*$) we will have gradients close to zero and the hessian matrix to be positive definite, i.e  $\nabla_{W_1,W_2}DIV = 0$ and $\nabla_{W_1,W_2}^2DIV \succ 0$.

\begin{thm} \cite{strang09}
A real symmetric matrix $H$ is positive definite if and only if its eigen values are positive.
\end{thm}

The hessian of a real valued function is real and symmetric. Thus, utilizing the above theorem we can verify the convergence of the neural network to a local minima of the objective function by ensuring that the eigenvalues of the hessian are all positive and the gradients are zero. 

\section{Experiments and Results \footnote{Code available at \url{https://github.com/sahilsid/svd-for-neural-networks}}}

\begin{figure*}[t]
\centering
\begin{tabular}{cccc}
\subfloat[IMDB 2D ]{\includegraphics[width = 0.20\linewidth]{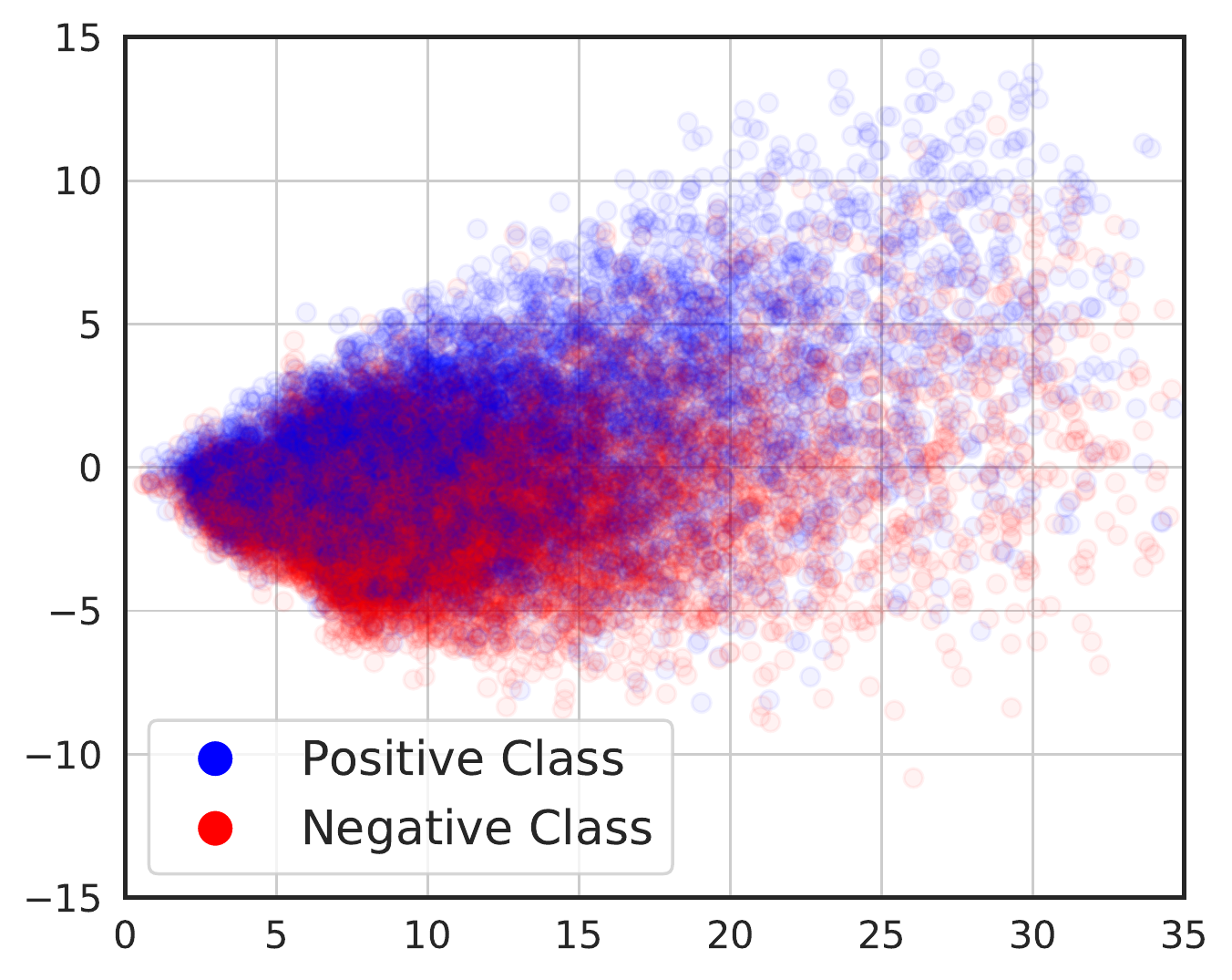}} &
\subfloat[IMDB 3D ]{\includegraphics[width = 0.25\linewidth]{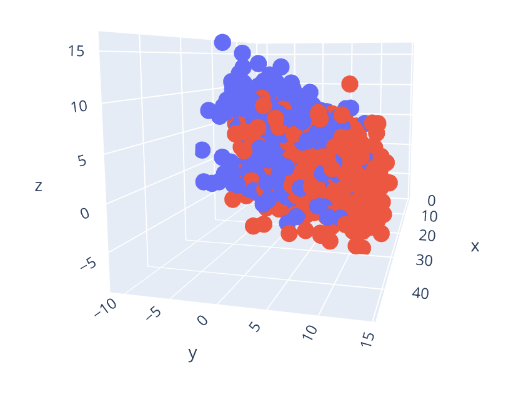}} &
\subfloat[Yelp 2D]{\includegraphics[width = 0.20\linewidth]{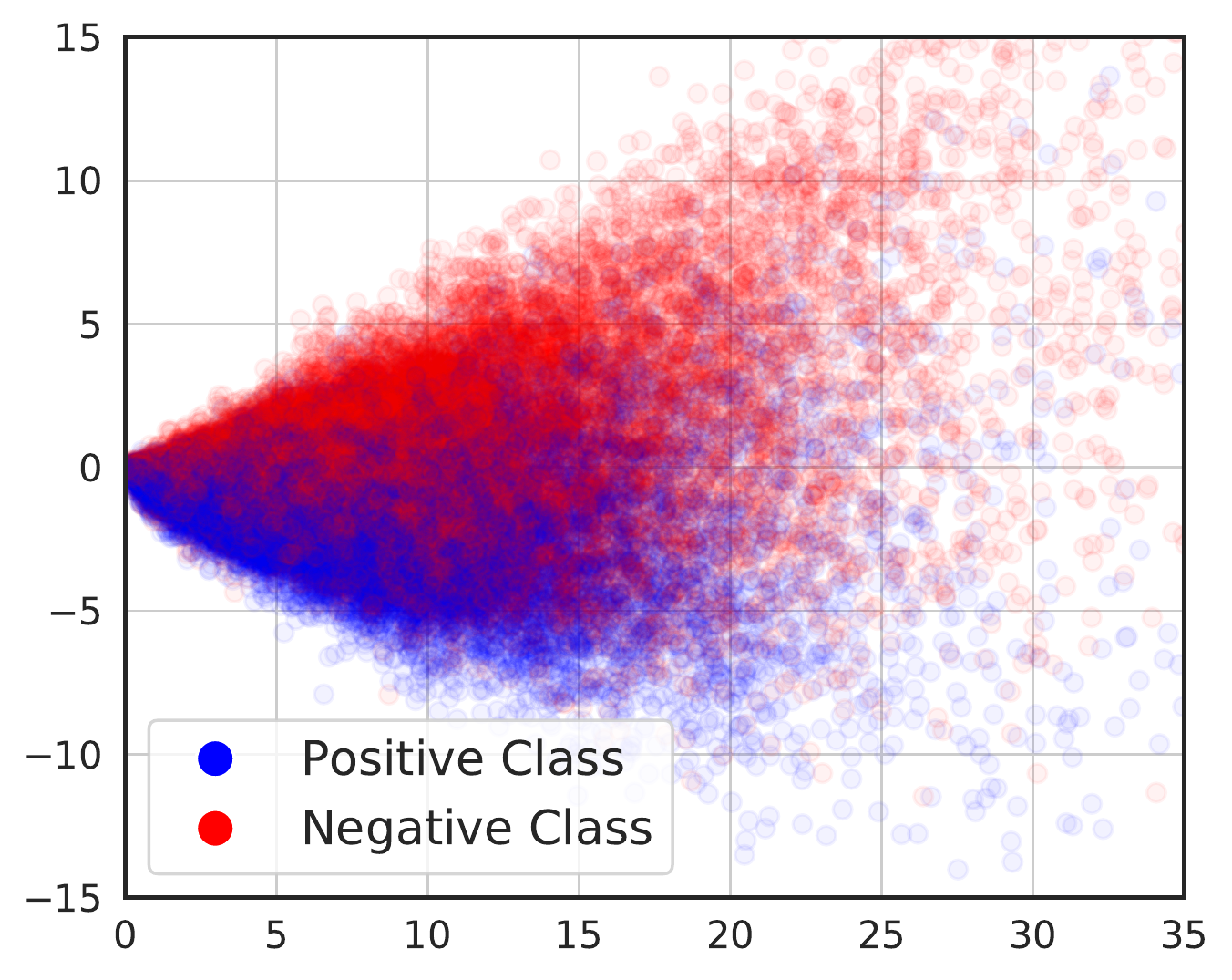}} &
\subfloat[Yelp 3D]{\includegraphics[width = 0.25\linewidth]{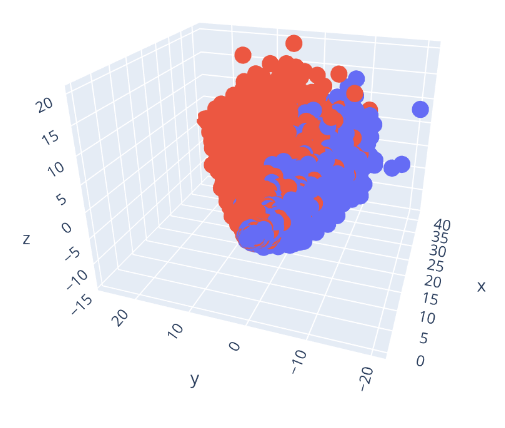}} \\

\\

\end{tabular}

\caption{Dataset Visualization by projecting to 2D and 3D space using SVD. Blue represents the positive class and red represents the negative class.}
\label{fig:dataset_visualization}
\end{figure*}

\begin{table*}
\normalsize
\centering
\begin{tabular}{@{}ccccccc@{}}
\toprule 
 & \multicolumn{4}{c}{ \textbf{Test Accuracy (\%)}} & \multicolumn{2}{c}{ \textbf{Computational Complexity}}  \\ 
\cmidrule(lr){2-5} \cmidrule(lr){6-7} 

  \multicolumn{1}{r}{ \textbf{}} & \multicolumn{2}{c}{IMDB} & \multicolumn{2}{c}{YELP}  \\ 
\cmidrule(lr){2-3} \cmidrule(lr){4-5}    
\multicolumn{1}{l}{\textbf{Latent Space Dim ($K$)}} & Binary & Count & Binary & Count  & \# Model Parameters & Ratio (\%) \\
\midrule
2                      & 53.244 & 54.095 & 66.302 & 57.326    & 1666 & 0.51\\ 
5                         & 64.476 & 59.808 &  69.665 & 64.952   & 2050 & 0.63  \\ 
10                     & 78.604 & 61.640 &  78.002 & 68.373     & 2690 & 0.83 \\ 
50                     & 79.963 & 73.396 & 83.971 & 81.200   & 7810 & 2.429 \\ 
100                     & 80.568 & 77.767 & 86.668 & 85.436  & 14210 & 4.421 \\ 
200                     & 82.219 & 80.328 &  88.810 & 88.239   & 27010 & 8.403\\ 
400                     & 84.876 & 83.424 & 90.236 & 89.997  & 52610 & 16.368\\ 
600                     & 85.344 & 84.752 & 90.197 & 90.544  & 78210 &  24.333 \\ 
800                    & 86.119 & 85.084 & 90.618 & 90.755 & 103810 & 32.298 \\ 
1000                     & 86.188 & 85.436 & 90.281 & 90.797   & 129410 & 40.263 \\ 
1500                    &\textbf{ 86.492 }& 85.976 & 90.678 & 91.218 & 193410 & 60.175 \\ 
2000  & 86.272 & \textbf{86.199} & \textbf{91.402} & \textbf{91.310}  & 257410 & 80.087\\ \addlinespace
\midrule
None                     & 85.692 & 81.104 & 91.263 & 90.476   & 321410 & 100.000  \\ 
 \addlinespace
\bottomrule

\\
\label{Tab:accuracy_table}
\end{tabular}
\\

\caption{Comparison of the performance and computational complexity of the neural network trained on SVD boosted latent spaces of varying dimesions ($K$). The bottom row represents the classical neural network trained on the original input space.}

\end{table*}

To study the effect of the latent space transformation on the model performance and computational complexity for text classification, we train a single hidden layer neural network on the following sentiment classification datasets:
\begin{itemize}
    \item \textbf{IMDB Movie Reviews} \cite{maas-EtAl:2011:ACL-HLT2011} Dataset comprising of 25000 training examples and 25000 testing examples, each of which are documents representing reviews for movies given by various users. The reviews are labelled positive or negative based on its sentiment. 
    
    \item \textbf{Yelp Polarity Reviews} \cite{zhang2016characterlevel} Dataset comprising of 560,000 training examples and 38000 testing examples, each of which are user reviews pertaining various businesses and services. The reviews are labelled positive or negative based on its polarity.
\end{itemize}

 For each of the datasets, we consider 2 different N-gram representations of the text data, namely \textit{binary} and \textit{count}. The \textit{binary} representation is a $0-1$ matrix indicating the presence of a term in a document, while the \textit{count} representation is an integer matrix depicting the number of times a term occurs in a document. We tokenize the text documents by defining a vocabulary of size 2500. Thus each document vector in the original input space belongs to $R^{2500}$.
 Our objective is not to compare the performance and complexity of sophisticated state of the art models in semantic classification tasks, but to show that even simple models such as a two layer feedforward network can benefit from significant reduction in the computational complexity without loss in performance when trained on the SVD boosted latent space.
 
\subsection{Dataset Visualization}

Figure \ref{fig:dataset_visualization} shows a visualization of each of the sentiment classification datasets in a 2-dimensional and 3-dimensional space obtained by projecting the datasets onto the corresponding dimensional latent space defined by the singular vectors of the document-term matrix. We can observe that the documents with positive and negative sentiments are to some extent clustered and distinguishable. However, there is a large amount of overlap between the classes in the depicted 2D and 3D latent space, indicating that accurately classifying the sentiments even on the latent space requires learning a capable model such as a neural network.

\subsection{Evaluating the effectiveness of the latent space }
 
\begin{figure*}
\centering
\begin{tabular}{cccc}
\subfloat[IMDB+$binary$]{\includegraphics[width = 0.235\linewidth]{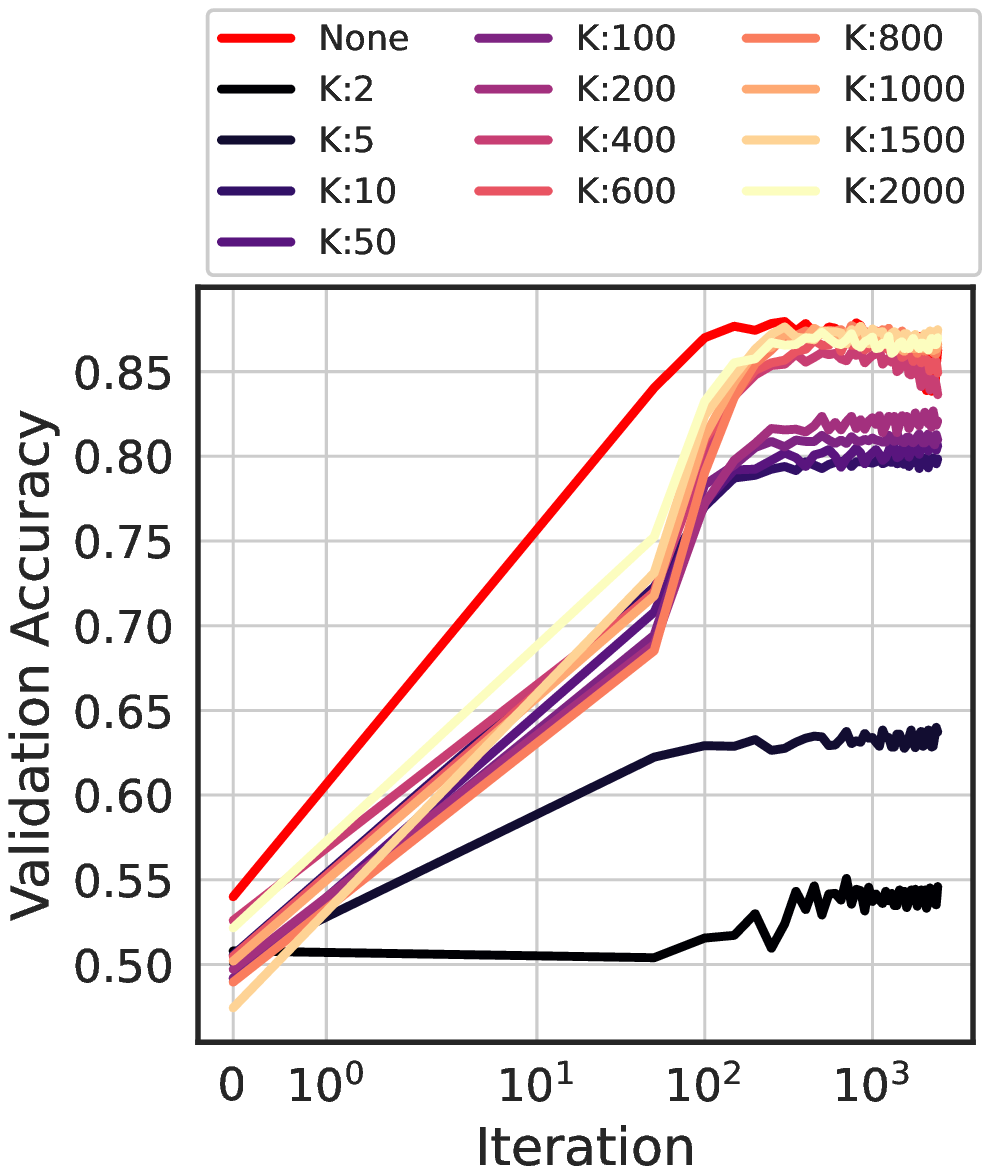}}&
\subfloat[IMDB+$count$]{\includegraphics[width = 0.235\linewidth]{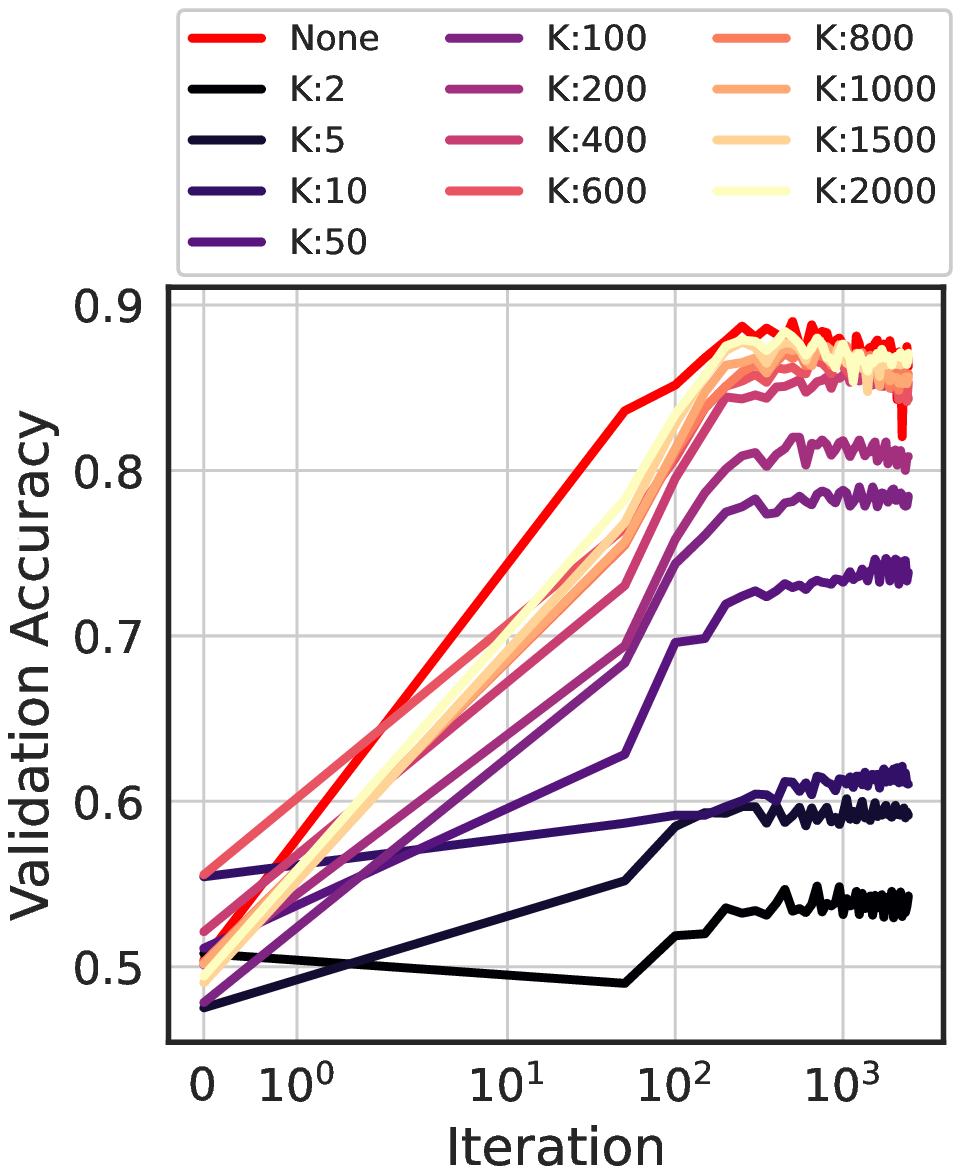}}&
\subfloat[Yelp+$binary$]{\includegraphics[width = 0.235\linewidth]{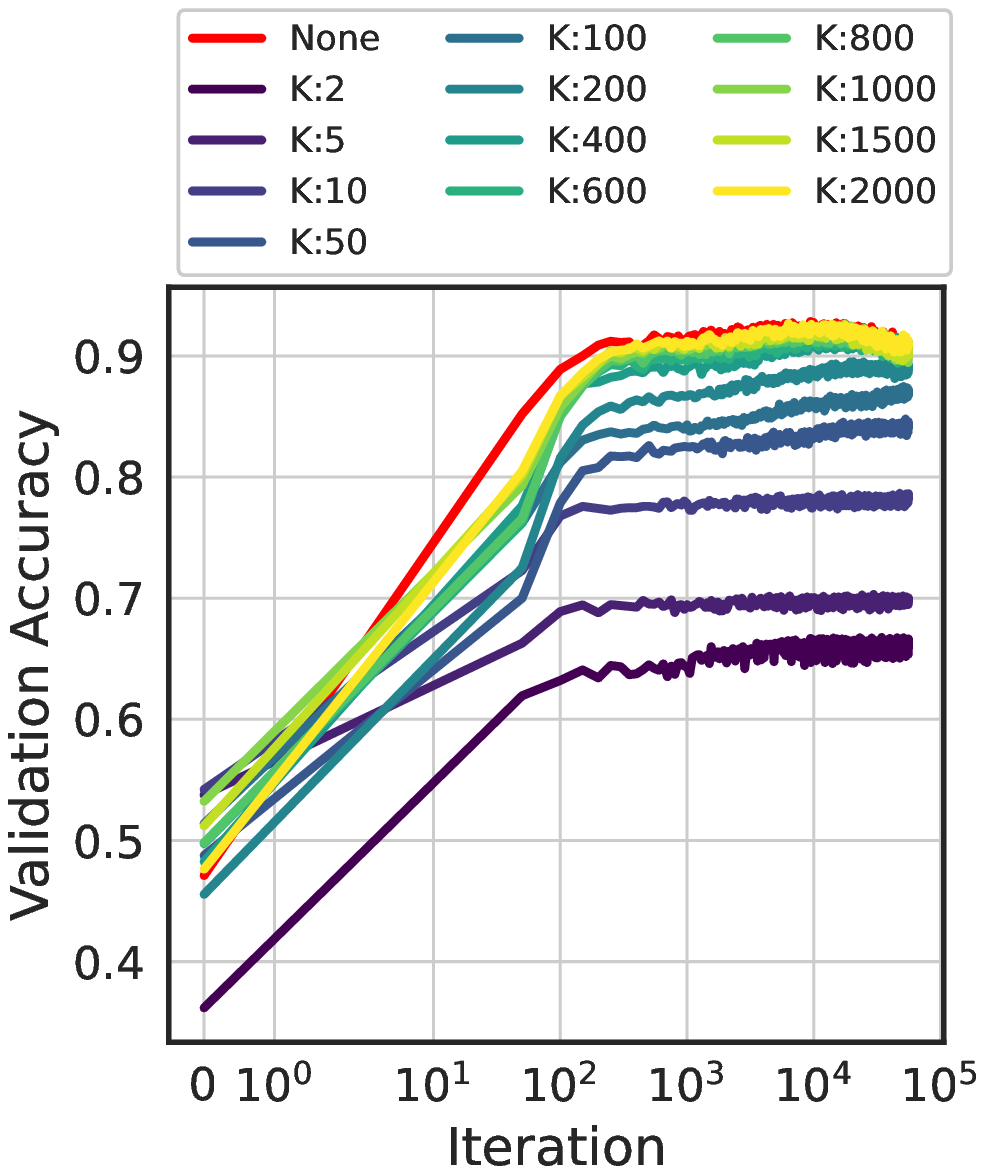}}&
\subfloat[Yelp+$count$]{\includegraphics[width = 0.235\linewidth]{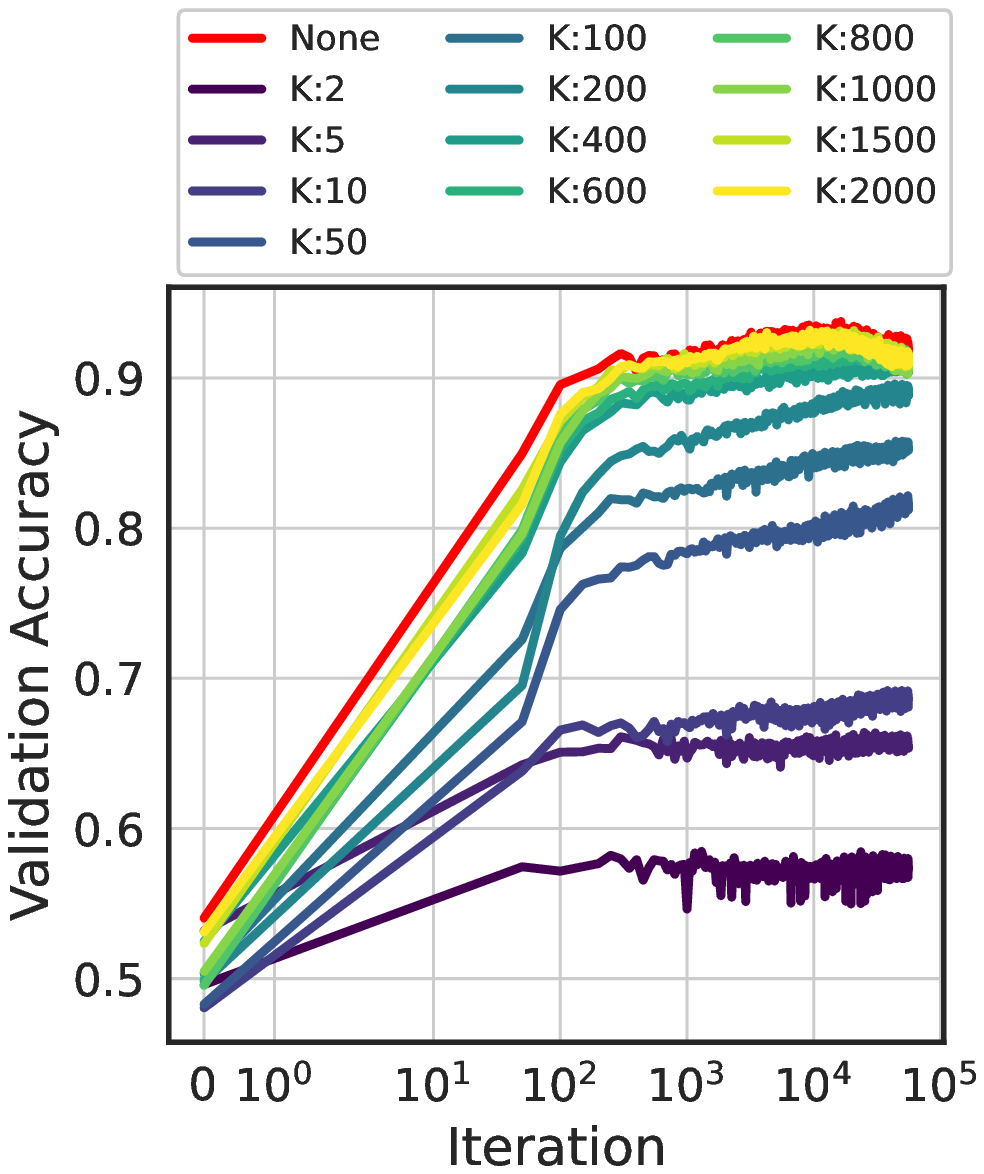}}\\

\\
\end{tabular}

\caption{Validation accuracy of neural networks trained using latent space transformation for varying dimension (K) across the two datasets and document matrix representations.}
\label{fig:validation_acc}
\end{figure*}

To study the efficacy of training the neural network on the latent space, we define latent spaces of varying dimensions($K)$ by considering the top-$K$ principal components and compare the performance of the neural network trained on this latent space as opposed to that trained on the original input space. Figure \ref{fig:validation_acc}
shows the trend of the validation accuracy of the neural network trained on the latent space for varying dimensions $K$. We observe that the learning quickly saturates to low accuracy values for very small values of $K$. On increasing $K$, there is an increase in the performance which further saturates beyond a particular $K'$ ($K'=600$ for IMDB and $K'=800$ for yelp). This performance is also at par with that given by the model trained on the higher dimensional input space. This suggests that a latent space of dimension is $K'$ is sufficient to capture all the information needed to perform the sentiment inference task.

\textbf{Table 1} summarizes the final test accuracies and the computational complexity in terms of the number of model parameters to be learned for each setting. The \% ratio of model parameters in terms of the original naive neural network trained on the input space is also given to understand the effectiveness of the latent space transformation in reducing the computational complexity. We can observe that training the neural network on the latent space helps achieve better test accuracies on all settings than training on the input space (bold values). This is expected because the dimensionality reduction retaining maximum information will remove the noise, sparsity and redundancy present in the data space, which in turn enables the neural network to easily learn the mapping to the target class space. A key observation here is that neural networks trained on the latent space require just 25\% of the model parameters to achieve performance within 1\% of the best performance as compared to the one trained on input space, thus significantly reducing the computational expense.

\subsection{Identifying the Optimal Latent Space Dimension}
\begin{figure}[t]
\centering
\begin{tabular}{cc}
\subfloat[IMDB]{\includegraphics[width = 0.47\linewidth]{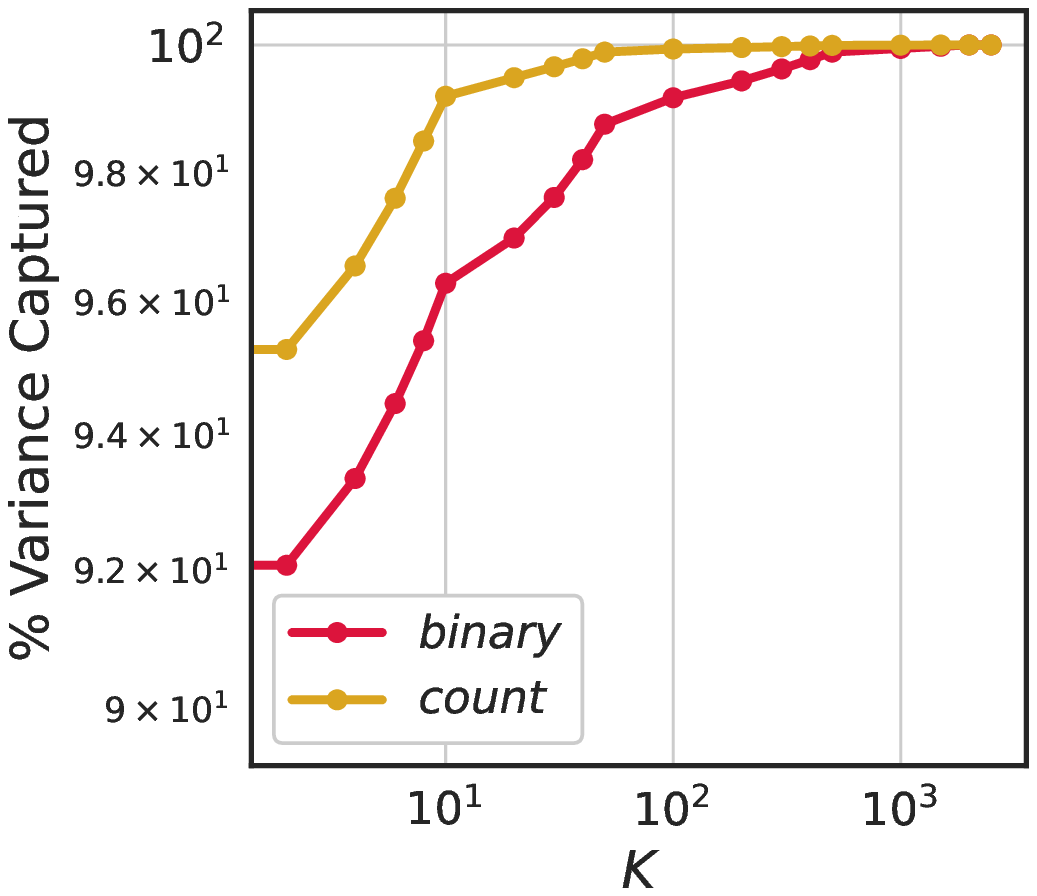}} &
\subfloat[Yelp]{\includegraphics[width = 0.47\linewidth]{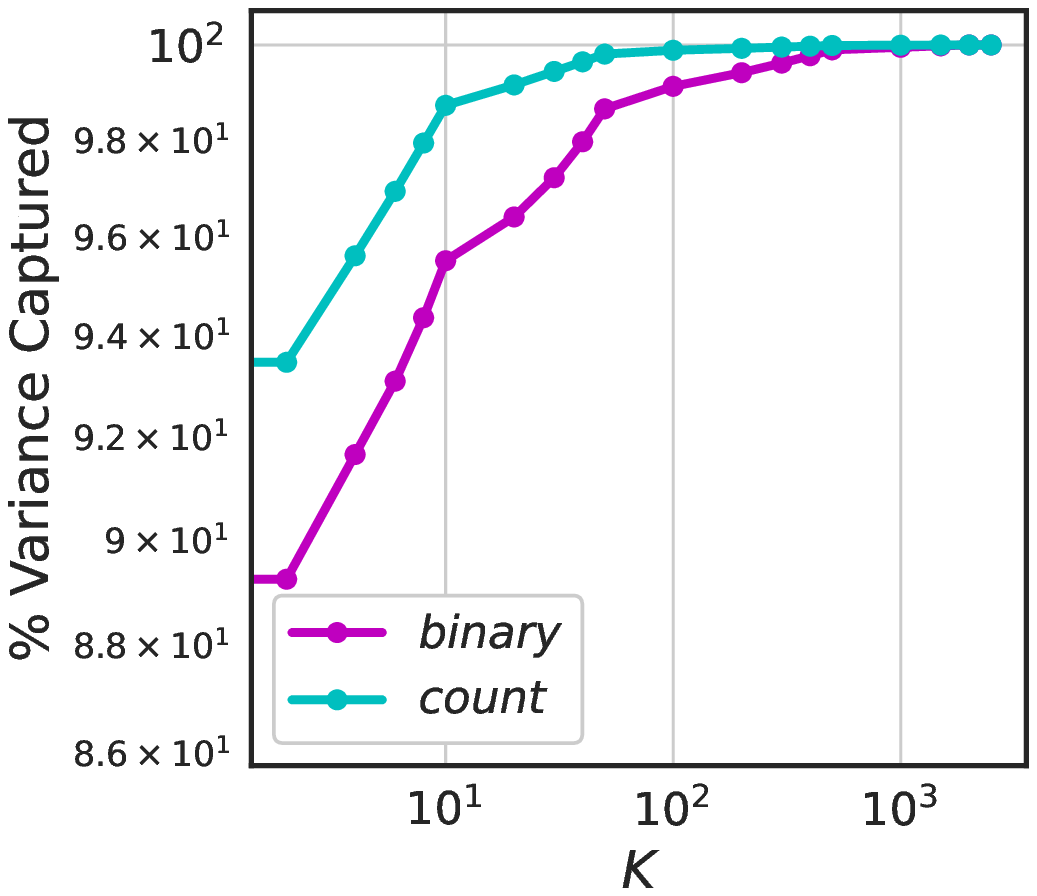}} 
\end{tabular}
\\
\caption{ Scree plot showing the percentage variance of the document matrix captured by latent spaces of varying dimensions ($K$) for the different representations of the datasets (a) IMDB and (b) Yelp.}
\label{fig:singular_values}
\end{figure}

The singular values of the document-term matrix is reflective of the variance in the data distribution along the directions defined by the corresponding singular vectors. To identify the optimal latent space dimension that de-noises the input data without loss of information, we plot the percentage variance of the data distribution that is captured within latent spaces of varying dimension $K$, also known as a scree plot. The percentage variance for a particular $K$ is computed as the ratio of the cumulative sum of squares of the singular values ( corresponding to the singular vectors) that define the $K$ dimensional subspace to the total sum of squares of all the singular values of the document-term matrix. We observe (figure \ref{fig:singular_values}) that for both the datasets, irrespective of the representation used (binary or count), the first 500 singular vectors ( i.e a latent space of dimension one-third of the input space) capture over 99\% of the total variance in the data distribution. The performance reported in table 1 for the corresponding latent spaces thus indicate that a latent space that retains over 99\% of variance in the data distribution can achieve significant performance gains while reducing computational expense. 
Thus the optimal latent space dimension can be estimated deterministically from the singular values of the document-term matrix, eluding the need for additional hyperparameter search.  

\subsection{Verification of Model Convergence} 

\begin{figure*}
\centering
\begin{tabular}{cccc}

\subfloat[IMDB+No SVD ]{\includegraphics[width = 0.23\linewidth]{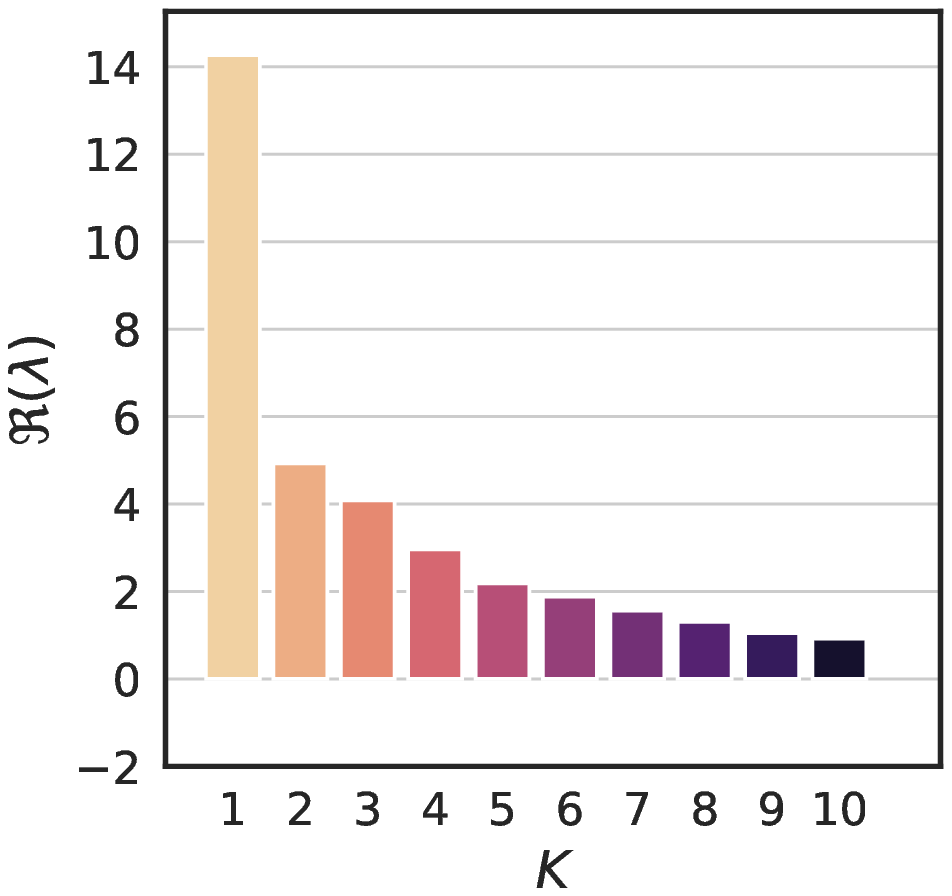}} &
\subfloat[IMDB+SVD ($K:600$)]{\includegraphics[width = 0.23\linewidth]{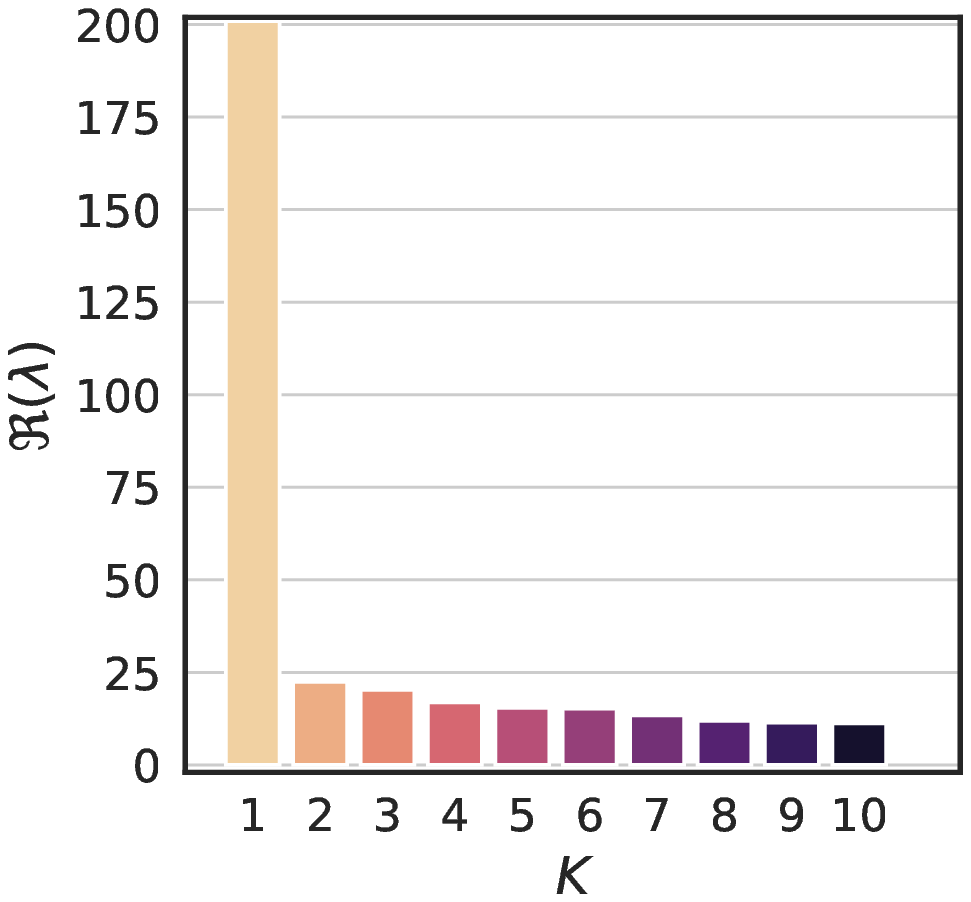}} &

\subfloat[Yelp + No SVD ]{\includegraphics[width = 0.23\linewidth]{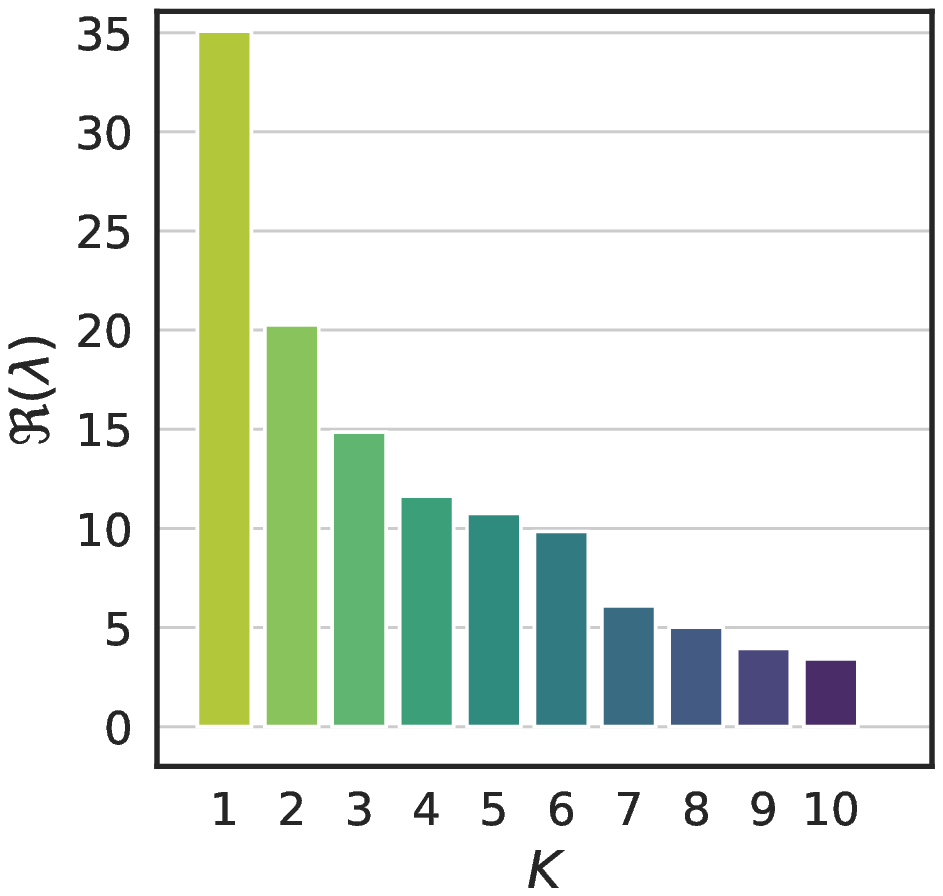}} &
\subfloat[Yelp + SVD ($K:600$)]{\includegraphics[width = 0.23\linewidth]{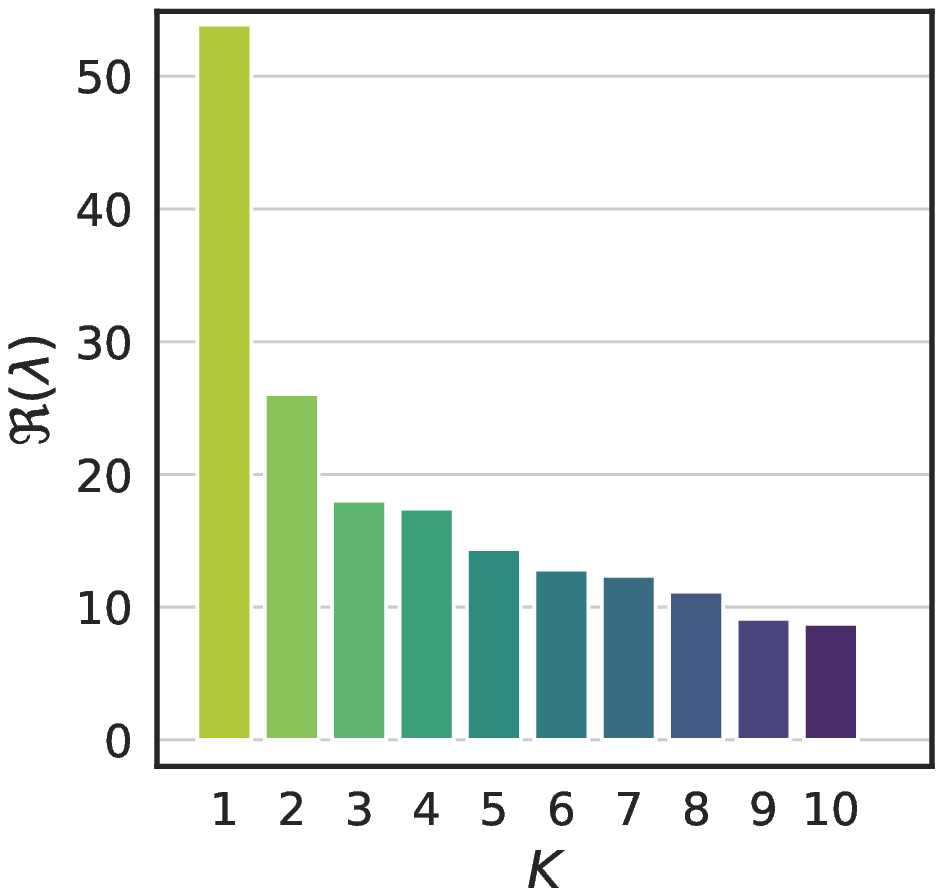}} \\
\\
\end{tabular}
\\
\caption{Top K eigen values of the hessian matrix when the neural network has converged are all positive, indicating it has converged to a local minima.}
\label{fig:eigen_values}
\end{figure*}

To ensure that the accuracies reported correspond to neural networks converged to the minima of the objective function, we compute the hessian matrix w.r.t the neural network parameters and find its eigen values. Ideally, at a minima, the hessian matrix should be positive definite, i.e. all its eigen values should be positive. Figure \ref{fig:eigen_values} shows the top-$K$ eigen values (in terms of magnitude) of the hessian matrix plotted for converged ( i.e gradient is zero) neural networks across different settings. We can observe that all the top-$K$ eigen values are positive, indicating that the neural networks have converged to a local minima.

\section{Conclusion}
Learning neural networks on noisy, sparse and high dimensional text data is computationally restrictive and slow. We can define better lower dimensional latent spaces for learning neural networks through singular value decomposition. Such a space results in an efficient input representation, that overcomes the above limitations and helps attain better downstream task performance. The reduction in the model complexity through utilizing the SVD boosted latent space also implicitly help in performing auxiliary analysis such as studying the convergence of neural network to a local minima through the computation of hessian matrices. Overall, this work celebrates the application of linear algebra to a machine learning domain, emphasizing that simple theories can have significant utility in day to day applications.

\bibliography{egbib}


\end{document}